\documentclass[conference]{IEEEtran}
\IEEEoverridecommandlockouts
\usepackage{cite}
\usepackage{amsmath,amssymb,amsfonts}
\usepackage{algorithmic}
\usepackage{graphicx}
\usepackage{textcomp}
\usepackage{xcolor}
\usepackage{makecell}
\usepackage{multirow}
\usepackage{pifont}
\usepackage{url}

\def\BibTeX{{\rm B\kern-.05em{\sc i\kern-.025em b}\kern-.08em
    T\kern-.1667em\lower.7ex\hbox{E}\kern-.125emX}}
\begin{document}

\title{\textit{Colo-SCRL}: Self-Supervised Contrastive Representation Learning for Colonoscopic Video Retrieval
\thanks{* indicates corresponding author. This work was partially supported by Wego Joint Lab.}
}

\makeatletter
\newcommand{\linebreakand}{%
  \end{@IEEEauthorhalign}
  \hfill\mbox{}\par
  \mbox{}\hfill\begin{@IEEEauthorhalign}
}
\makeatother

\author{\IEEEauthorblockN{Qingzhong Chen}
\IEEEauthorblockA{\textit{School of Biomedical Engineering} \\
\textit{Shanghai Jiao Tong University}\\
Shanghai, China \\
chenqz1998@sjtu.edu.cn}
\and
\IEEEauthorblockN{Shilun Cai}
\IEEEauthorblockA{\textit{Endoscopy Center} \\
\textit{Zhongshan Hospital of Fudan University}\\
Shanghai, China \\
caishilun1988@qq.com}
\and
\IEEEauthorblockN{Crystal Cai}
\IEEEauthorblockA{\textit{School of Biomedical Engineering} \\
\textit{Shanghai Jiao Tong University}\\
Shanghai, China \\
crystal.cai@sjtu.edu.cn}
\linebreakand
\IEEEauthorblockN{Zefang Yu}
\IEEEauthorblockA{\textit{School of Electronic Information} \\
\textit{and Electrical Engineering}\\
\textit{Shanghai Jiao Tong University}\\
Shanghai, China \\
yuzefang@sjtu.edu.cn}
\and
\IEEEauthorblockN{Dahong Qian$^{*}$}
\IEEEauthorblockA{\textit{School of Biomedical Engineering} \\
\textit{Shanghai Jiao Tong University}\\
Shanghai, China \\
dahong.qian@sjtu.edu.cn}
\and
\IEEEauthorblockN{Suncheng Xiang$^{*}$}
\IEEEauthorblockA{\textit{School of Biomedical Engineering} \\
\textit{Shanghai Jiao Tong University}\\
Shanghai, China \\
xiangsuncheng17@sjtu.edu.cn}
}
\maketitle

\begin{abstract}
Colonoscopic video retrieval, which is a critical part of polyp treatment, has great clinical significance for the prevention and treatment of colorectal cancer.
However, retrieval models trained on action recognition datasets usually produce unsatisfactory retrieval results on colonoscopic datasets due to the large domain gap between them.
To seek a solution to this problem, we construct a large-scale colonoscopic dataset named \textbf{Colo-Pair} for medical practice. Based on this dataset, a simple yet effective training method called \textbf{\textit{Colo-SCRL}} is proposed for more robust representation learning. It aims to refine general knowledge from colonoscopies through masked autoencoder-based reconstruction and momentum contrast to improve retrieval performance. To the best of our knowledge, this is the first attempt to employ the contrastive learning paradigm for medical video retrieval. Empirical results show that our method significantly outperforms current state-of-the-art methods in the colonoscopic video retrieval task.
\end{abstract}


\section{Introduction}
Colorectal cancer (CRC) is the second leading cause of cancer death globally with an estimated 151,030 new cases and 52,580 deaths in 2022~\cite{2022Cancer}. As the gold standard for CRC screening, colonoscopies can significantly reduce the risk of death from CRC by early detection of tumors and removal of precancerous lesions~\cite{xu2022deep}. For polyps found in previous colonoscopies, it usually takes time and tremendous efforts to locate and match them during colonoscopic polyp treatment. The repeated operation to find such polyps not only delays the treatment time, but also easily leads to patients developing abdominal distension, abdominal pain and other complications. Essentially, localization failure is a missed diagnosis of known polyps. Therefore, timely and accurate localization of polyps through colonoscopies is of great clinical significance for the prevention and treatment of CRC.
\begin{figure*}[t]
  \centering
  \centerline{\includegraphics[width=1.0\linewidth]{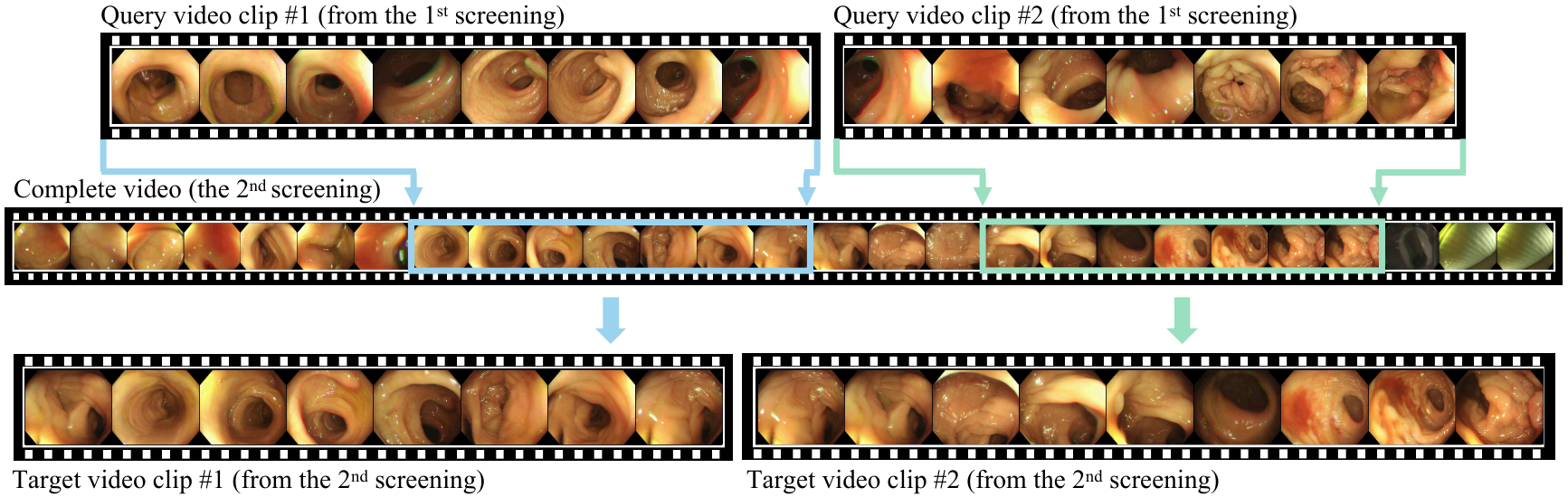}}
  \caption{Illustration of colonoscopic video retrieval. The top query video clips are polyp areas captured from the first colonoscopy screening. The middle video is the complete second colonoscopy screening which contains the corresponding target video clips, marked by the rectangles and shown in the bottom.}
  \label{fig1}
\end{figure*}

As illustrated in Fig.~\ref{fig1}, given the query video clip, colonoscopic video retrieval aims to accurately locate the similar clip which semantically corresponds to the given query one.
Up until now, no solution to this problem has been proposed in the deep learning community. Inspired by computer vision technology~\cite{kordopatis2019visil, shao2021temporal, kordopatis2022dns, xiang2020unsupervised, xiang2021less}, we take a big step forward and develop a colonoscopic video retrieval approach to solve the aforementioned problems. Specifically, we capture polyp areas from the first colonoscopy screening and treat them as the query video clips to be retrieved. We then retrieve video clips of matched polyp areas from the second colonoscopy screening. When the polyp areas are successfully matched, physicians are alerted to pay attention to the possible presence of previously found polyps in the nearby area.
Unfortunately, high similarity of the inner colon environment, existence of numerous colon folds, peristalsis of the colon itself, and blind areas in colonoscopy all bring significant challenges for colonoscopic video retrieval. Additionally, data collection and labeling are time-consuming for pairwise polyp area data. Data scarcity also makes it difficult for methods that rely on a sufficient amount of data to work.

To address these challenges, we present the \textbf{Colo-Pair} dataset, a large-scale manually-constructed dataset of complete colonoscopy sequences acquired during regular medical practice, including slow and careful screening explorations.
On top of this, a simple but effective \textbf{S}elf-supervised \textbf{C}ontrastive \textbf{R}epresentation \textbf{L}earning framework named \textbf{\textit{Colo-SCRL}} is proposed to refine general knowledge from our colonoscopies. To be more specific, for the pre-training process, we employ the video masked autoencoder (VideoMAE), which inherits the merits of masking random cubes in the video and reconstructing the missing ones, to encourage extracting more effective video representations. For the downstream retrieval process, we build a contrastive video representation learning framework to learn spatial representations from unlabeled videos, which adopts InfoNCE contrastive loss to compare the similarity between positive and negative pairs. To the best of our knowledge, this is the first attempt to apply the contrast learning paradigm to medical video retrieval. We hope our dataset and method will shed light into future directions for the medical deep learning community.

Thus, our main contributions are summarized as follows:
\begin{itemize}
  \item We construct the first colonoscopy video dataset with pairs of complete screenings from each patient to facilitate the development for medical practice.
  \item Based on this dataset, a self-supervised contrastive representation learning scheme \textbf{\textit{Colo-SCRL}} is proposed to learn spatial representations from our video dataset on the colonoscopic video retrieval task.
  \item Extensive experiments on benchmarks demonstrate the superiority of our proposed framework compared with current state-of-the-art methods.
\end{itemize}

\section{Related Works}

\subsection{Content-Based Video Retrieval}

The research field of content-based video retrieval has gone through rapid development in recent years~\cite{kordopatis2019visil, shao2021temporal, kordopatis2022dns}. For example, Kordopatis-Zilos \textit{et al.} introduced a video similarity learning architecture named ViSiL\cite{kordopatis2019visil} and DnS~\cite{kordopatis2022dns} for video similarity learning. Shao \textit{et al.}~\cite{shao2021temporal} proposed Temporal Context Aggregation for content-based video retrieval. However, these solutions are incapable of capturing a large variety of temporal similarity patterns due to their rigid aggregation approach, which hinders the further improvement for colonoscopic video retrieval tasks.

\subsection{Self-Supervised Video Representation Learning}

Self-supervised video representation learning has recently witnessed rapid progress in image classification, segmentation and object detection~\cite{he2020momentum, han2020self, tong2022videomae, pan2021videomoco, wang2021self, qian2021spatiotemporal}. Inspired from this, investigations on video representation learning have now attracted enormous attention from both academia and industry. For example, He \textit{et al.}~\cite{he2020momentum} presented momentum contrast for unsupervised visual representation learning. Han \textit{et al}.~\cite{han2020self} proposed a novel self-supervised co-training scheme to improve the popular InfoNCE loss. Tong~\textit{et al.}~\cite{tong2022videomae} proposed an effective VideoMAE that unleashes the potential of vanilla vision transformer for video recognition. VideoMoCo, designed by Pan~\textit{et al.}~\cite{pan2021videomoco} for self-supervised video representation learning, delves into MoCo and empowers its temporal representation by introducing temporally adversarial learning and temporal decay. Wang~\textit{et al.}~\cite{wang2021self} introduced a new task for self-supervised video representation learning by uncovering a set of spatio-temporal labels derived from appearance statistics. Although these methods have promoted the accuracy to a new degree, the performance of video representation learning is still unsatisfactory since previous approaches fail to explore the potential of the contrastive learning paradigm.

\section{The Colo-Pair Dataset}

As illustrated in Fig.~\ref{fig2}, we manually construct the Colo-Pair dataset in this work. It is the first collection of complete paired colonoscopy sequences acquired during regular medical practice, \textit{i.e.,} slow and careful screening explorations.

Specifically, the colonoscopy videos were collected from patients undergoing routine colonoscopy at Zhongshan Hospital of Fudan University\footnote{\url{https://www.zs-hospital.sh.cn/}}. An Olympus CV-290 colonoscope with its built-in light source and image processor was used. Videos were recorded at 1920 $\times$ 1080 resolution with 25 frames per second. The dataset consists of 60 videos from 30 patients, with 62 query video clips and the corresponding polyp clips from the second screening manually annotated as positive retrieval clips. We collected in total 9.6 hours of paired videos with full procedures, including two colonoscopy screenings for each patient, designed to facilitate the development and evaluation of polyp area location in real colonoscopy data. Additionally, we also provide labels of anatomical landmarks such as the ileocecal valve, the hepatic flexure, and the splenic flexure. These landmarks can not only act as prior knowledge during video retrieval but also promises to promote the development of other tasks, \textit{e.g.}, colon segment classification~\cite{yao2021motion}.

We compare our constructed dataset with publicly available datasets in Table \ref{tab:dataset}, which shows that no other public dataset offers a comparable volume of full colonoscopies.
More importantly, unlike other public datasets which provide only single colonoscopic screenings of each patient, we collect pairs of colonoscopy videos with semantic correspondence and manually label the corresponding clips, which satisfies the need for the retrieval task.

\begin{figure}[!t]
  \centering
  \centerline{\includegraphics[width=1.0\linewidth]{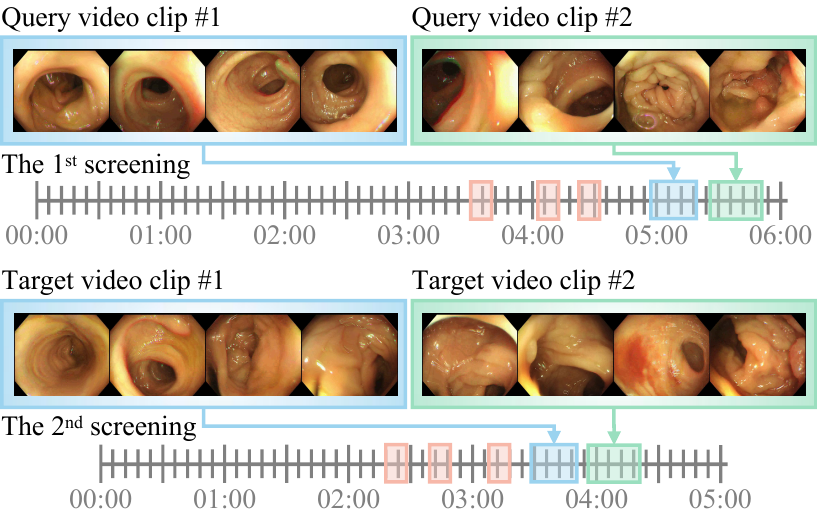}}
  \caption{An example of our colonoscopic dataset Colo-Pair. The top and bottom videos make up a pair of complete colonoscopic screenings from the same patient. The \textcolor[rgb]{0.125,0.578,0.805}{blue} and \textcolor[rgb]{0.203,0.641,0.441}{green} rectangles represent polyp areas to be retrieved, and the \textcolor[rgb]{0.910,0.270,0.125}{red} ones represent anatomical landmarks, \textit{i.e.,} ileocecal valve, hepatic flexure and splenic flexure, which act as prior knowledge to narrow the scope of video retrieval.}
  \label{fig2}
\end{figure}

\begin{table}[b]
\begin{center}
\caption{Overview of existing endoscopy datasets.} \label{tab:dataset}
\begin{tabular}{|l|l|l|}
\hline
\textbf{Dataset} & \textbf{Type of Data} & \textbf{Size of Dataset} \\ \hline
Artefacts~\cite{ali2020objective} & Images & 5138 images \\ \hline
GIANA 2021~\cite{bernal2021polyp} & \begin{tabular}[c]{@{}l@{}}Images \&\\ short videos\end{tabular} & \begin{tabular}[c]{@{}l@{}}38 videos \&\\ 3000 images\end{tabular} \\ \hline
Kvasir~\cite{pogorelov2017kvasir} & Images & 4000 images \\ \hline
Kvasir-Seg~\cite{jha2020kvasir} & Images & 8000 images \\ \hline
Nerthus~\cite{pogorelov2017nerthus} & Short videos & 21 videos \\ \hline
Heilderberg~\cite{maier2021heidelberg} & Images & 10040 images \\ \hline
HyperKvasir~\cite{borgli2020hyperkvasir} & \begin{tabular}[c]{@{}l@{}}Images \&\\ short videos\end{tabular} & \begin{tabular}[c]{@{}l@{}}110079 images \&\\ 374 short videos\end{tabular} \\ \hline
Endomapper~\cite{azagra2022endomapper} & Videos & \begin{tabular}[c]{@{}l@{}}20 videos\\ ($\sim$7.5 hours)\end{tabular} \\ \hline
\textbf{Colo-Pair (Ours)} & \textbf{Paired videos} & \textbf{\begin{tabular}[c]{@{}l@{}}60 videos\\ ($\sim$9.6 hours)\end{tabular}} \\ \hline
\end{tabular}
\end{center}
\end{table}

\section{The Colo-SCRL Framework}

\subsection{Overview}
As illustrated in Fig. \ref{fig3}, the training process of \textit{Colo-SCRL} can be described in two steps. First, during the pre-training stage, the VideoMAE is employed to refine general colonoscopic knowledge through masking and reconstruction tasks and the learned encoder is transferred to downstream task. Second, during the downstream colonoscopic video retrieval stage, the core of contrastive video representation learning is an InfoNCE contrastive loss applied on features extracted from augmented videos.

\begin{figure*}[!t]
  \centering
  \centerline{\includegraphics[width=1\linewidth]{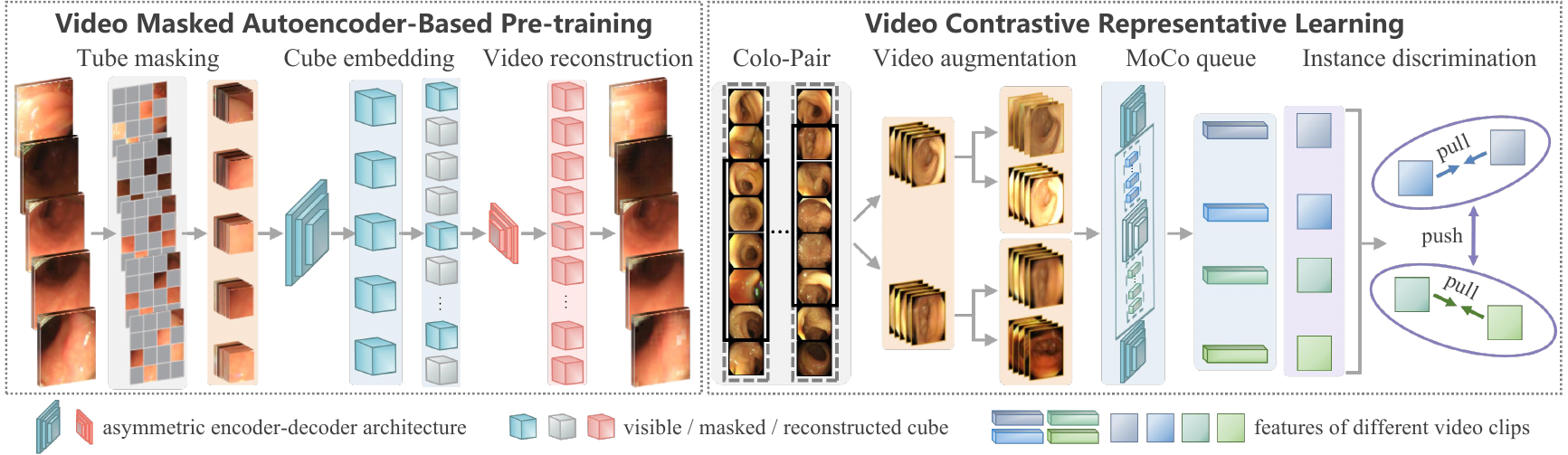}}
  \caption{Overview of the proposed \textit{Colo-SCRL} framework. In the pre-training stage, we propose the use of VideoMAE and introduce the strategy of tube masking to effectively reduce the risk of time-dependent information leakage during video reconstruction. In the downstream colonoscopic video retrieval stage, we adopt the instance discrimination strategy to fine-tune the pre-trained encoder in an unsupervised way. MoCo-based learning is introduced to build a large and consistent dictionary, thus enhancing the discriminant ability of the model.}
  \label{fig3}
\end{figure*}

\subsection{Video Masked Autoencoder-Based Pre-training}

We introduce a simple and data-efficient self-supervised learning method called VideoMAE for video transformer pre-training. The core of VideoMAE is to perform the masking and reconstruction tasks using an asymmetric encoder-decoder architecture~\cite{tong2022videomae}. Firstly, the joint space-time cube embedding is adopted in VideoMAE, where each input video clip $V\in \mathcal{R} ^{Frame\times 3\times H\times W}$ is divided into $2\times 16\times16$ regular non-overlapping cubes as one token embedding. Secondly, these tokens are masked randomly at a high masking rate (90\%), and only the remaining tokens are fed into the transformer encoder. Finally, a shallow decoder is placed on top of the visible tokens from the encoder and learnable mask tokens to reconstruct the video clip. We employ mean square error (MSE) loss as the loss function between the normalized masked tokens and the reconstructed ones in pixel level:
\begin{eqnarray}
L_{MSE}=\frac{1}{\Omega } \sum_{p\in \Omega } \left | V(p)-\hat{V}(p)  \right | ^2
\label{videomae_loss}
\end{eqnarray}
where $p$ is the token index, $\Omega$ is the masked token set, $V$ is the input video clip and $\hat{V}$ is the reconstructed one.

We utilize the temporal tube masking mechanism to improve the masking efficiency by alleviating information leakage due to temporal redundancy~\cite{tong2022videomae}. Temporal tube masking forces a mask to extend across the whole timeline, \textit{i.e.}, different frames share the same masking map and temporal neighbors of masked cubes are also masked. Such mechanism encourages VideoMAE to utilize high-level semantics to recover missing space-time cubes and makes masked video reconstruction a meaningful self-supervised pre-training task.

\subsection{Video Contrastive Representation Learning}

\noindent\textbf{Momentum Contrast.} We introduce the Momentum Contrast (MoCo) strategy to build a large and consistent dictionary to cache a large number of features for unsupervised learning. It views the process of querying positive samples in contrast learning as a dictionary look-up task~\cite{he2020momentum}. Given a dictionary queue including an encoded query \textit{q} and encoded keys \{\textit{$k_0$}, \textit{$k_1$}, \textit{$k_2$},...\}, the contrastive loss function of MoCo can then be written as:
\begin{eqnarray}
L_q =-log\frac{exp(q\cdot k_+/\tau)}{ {\textstyle \sum_{i=0}^{K}exp(q\cdot k_i/\tau)} }
\label{moco_loss}
\end{eqnarray}

\noindent where similarity is measured by dot product and $\tau$ is a temperature parameter. The sum is over one positive and \textit{K} negative samples. This loss function tries to classify \textit{q} as \textit{$k_+$} in a softmax-based classifier. The query \textit{q} denotes a representation of the input sample obtained by the encoder, and the keys \textit{$k_i$} denote representations of the other training samples in the queue. We consider \textit{q} and \textit{$k_i$} as a positive pair if they originate from the same video clip and otherwise as a negative pair.

The critical part of MoCo is the dynamically-maintained and momentum-updated queue. The current mini-batch is enqueued and the oldest mini-batch is dequeued, following the FIFO (first in first out) scheme. Thus, dictionary size is decoupled from the mini-batch size. We denote the parameters of the encoder as $\theta_q$ and the parameters of the momentum encoder as $\theta_k$. Only the encoder is updated by back-propagation with the computed contrastive loss in (\ref{moco_loss}), while the momentum encoder is updated more slowly based on the encoder change as:
\begin{eqnarray}
\theta _k\gets m*\theta _k+(1-m)*\theta _q
\end{eqnarray}

\noindent where \textit{m} $\in$ [0,1) is a momentum coefficient. A large momentum (\textit{e.g.}, $m = 0.999$) ensures a stable key representation and works better than a relatively small one (\textit{e.g.}, $m = 0.9$)~\cite{he2020momentum}. It suggests that the slowly updated key encoder makes good use of the queue. Thus, though the keys in the queue are encoded by different encoders (in different mini-batches), the differences between these encoders can be small.

\noindent\textbf{Instance Discrimination.} In self-supervised contrastive video representation learning, given a dataset $D$ containing $N$ raw video clips, the objective is to obtain a function $f(\cdot )$ to encode video clips for different downstream tasks, such as colonoscopic video retrieval.

Suppose there is a video augmentation function $\psi (\cdot ;a)$, where $a$ is collected from a set of pre-defined data augmentation transformations $A$ (applied to $D$). For a sample $x_i$, its positive sample set $P_i$ and negative sample set $N_i$ are defined as $P_i=\{\psi (x_i;a)|a\sim A\}$, and $N_i=\{\psi (x_n;a)|\forall n\ne i,a\sim A\}$. Given $z_i=f(\psi(x_i;\cdot ))$, the contrastive loss function (InfoNCE~\cite{oord2018representation}) in (\ref{moco_loss}) can be specifically written as:
\begin{eqnarray}
L_q =-log\frac{exp(z_i\cdot z_p/\tau)}{exp(z_i\cdot z_p/\tau)+ {\textstyle \sum_{n\in N_i}exp(z_i\cdot z_n/\tau )} }
\label{infonce_loss}
\end{eqnarray}

\noindent which allows the positive pair $(z_i,z_p)$ to attract mutually while the negative pair $(z_i,z_n)$ repels each other. In essence, the optimization process can be regarded as instance discrimination~\cite{wu2018unsupervised} tasks, \textit{i.e.}, the augmented views of the same instance get higher similarity scores than those of other different instances.

\noindent\textbf{Video Augmentation.}
Two random views of the same video clip are generated by employing random video augmentations and will exemplify a positive pair. A simple strategy is to use existing image-based data augmentation methods (\textit{e.g.}, random rotation, random cropping, color jittering and blurring) to each video frame. However, this strategy may negatively affect representation learning along the temporal dimension in videos~\cite{qian2021spatiotemporal}. Therefore, we make image augmentations consistent along the temporal dimension, \textit{i.e.}, hyper-parameters are generated only once per video clip and applied to all frames.

\section{Experiments}

\subsection{Experimental Settings}

\noindent\textbf{Implementation Details.}
All models are implemented in the Pytorch library with NVIDIA GeForce RTX 2080Ti GPUs. We choose the vanilla ViT~\cite{caron2021vit} architecture as the feature extractor. At the pre-training stage, the parameter settings follow the practice of VideoMAE~\cite{tong2022videomae}. At the contrastive learning stage, we train the network with InfoNCE for 500 epochs, where an epoch means to have sampled every clip from each video. For optimization, we use Adam with $10^{-3}$ learning rate and $10^{-5}$ weight decay. The learning rate is decayed down by $1/10$ twice when the validation loss plateaus. For data augmentation, we apply random cropping, horizontal flipping, Gaussian blur and color jittering, which are clip-wise consistent. It is worth mentioning that both non-polyp region and polyp region video clips are used in the training stage, while the query and target video clips consist mainly of portions in which polyp regions are present in the evaluation stage.

\noindent\textbf{Evaluation.}
We verify the effectiveness of \textit{Colo-SCRL} and compare its performance with other methods on our Colo-Pair dataset.
The standard evaluation metrics are employed to evaluate the performance of retrieval, including the standard rank at $k$ (R@$k$) and mean Average Precision (mAP). R@$k$ is defined as the fraction of queries for which the correct items belong to the top-$k$ ($k\in [1, 5, 10]$) retrieved items, and mAP is the average of all precision values computed when each relevant target is retrieved.

\subsection{Comparison with Other Methods}
In this section, we compare our proposed \textit{Colo-SCRL} with eight state-of-the-art content-based video retrieval (CBVR) methods as shown in Table \ref{tab:comparison}. Among them, ViSiL, CgS$^c$, FgAttS$^f_A$ and FgBinS$^f_B$ are pre-trained on the FIVR-200K or DnS-100K dataset, and CoCLR and ViT are pre-trained on the UCF-101 dataset, considering that these methods fail to work when the number of positive samples for training is small. TCA and CVRL are trained on the Colo-Pair dataset. Benefiting from its automatic hard negative mining and the memory bank mechanism, TCA achieves the second-best results next to \textit{Colo-SCRL}. However, the sampled keys in the memory bank for TCA are less consistent~\cite{he2020momentum}. Our method explores the potential of MoCo to increase the capacity of negative samples, which allows it to be more memory-efficient and adaptable in real scenarios.

As shown in Table \ref{tab:comparison}, our method significantly outperforms the other eight methods, demonstrating the superiority of our self-supervised contrastive representation learning scheme. On the Colo-Pair dataset, our method achieves R@1=22.6\%, R@5=41.9\%, R@10=58.1\% and mAP=0.315, showing \textbf{+6.5\%}, \textbf{+6.4\%}, \textbf{+4.9\%} and \textbf{+0.037} relative improvement compared to the current state-of-the-art CBVR method, respectively.

\begin{table}[b]
\begin{center}
\caption{Performance comparison with other methods (R@$k$: \%) on our Colo-Pair Dataset. Best results are marked in \textbf{bold}.} \label{tab:comparison}
\begin{tabular}{lccccc}
\Xhline{0.8pt}
\multirow{2}{*}{Method} & \multirow{2}{*}{Venue} & \multicolumn{4}{c}{Video Retrieval $\uparrow$} \\ \cline{3-6}
 &  & R@1 & R@5 & R@10 & mAP \\ \hline
ViSiL~\cite{kordopatis2019visil} & ICCV 19 & 14.5 & 30.6 & 51.6 & 0.249 \\
CoCLR~\cite{han2020self} & NIPS 20 & 6.5 & 22.6 & 33.9 & 0.163 \\
TCA~\cite{shao2021temporal} & WCAV 21 & 16.1 & 35.5 & 53.2 & 0.278 \\
ViT~\cite{caron2021vit} & CVPR 21 & 9.7 & 30.6 & 43.5 & 0.204 \\
CVRL~\cite{qian2021spatiotemporal} & CVPR 21 & 11.3 & 32.3 & 53.2 & 0.236 \\
CgS$^c$~\cite{kordopatis2022dns} & IJCV 22 & 8.1 & 35.5 & 45.2 & 0.214 \\
FgAttS$^f_A$~\cite{kordopatis2022dns} & IJCV 22 & 9.7 & 40.3 & 50.0 & 0.236 \\
FgBinS$^f_B$~\cite{kordopatis2022dns} & IJCV 22 & 9.7 & 32.3 & 48.4 & 0.212 \\
\hline
\textbf{\textit{Colo-SCRL}} & \textbf{Ours} & \textbf{22.6} & \textbf{41.9} & \textbf{58.1} & \textbf{0.315} \\
\Xhline{0.8pt}
\end{tabular}
\end{center}
\end{table}

\begin{table}[!t]
\begin{center}
\caption{Ablation results on our \textit{Colo-SCRL} modules (R@$k$: \%). ``BL", ``MAE" and ``CRL" represent the baseline model, VideoMAE and contrastive representation learning, respectively.  Best results are marked in \textbf{bold}.} \label{tab:ablation}
\setlength{\tabcolsep}{2.45mm}{
\begin{tabular}{ccccccc}
\Xhline{0.8pt}
\multirow{2}{*}{BL} & \multirow{2}{*}{MAE} & \multirow{2}{*}{CRL} & \multicolumn{4}{c}{Video Retrieval $\uparrow$} \\ \cline{4-7}
 &  &  & R@1 & R@5 & R@10 & mAP \\ \hline
\ding{51} & \ding{55} & \ding{55} & 9.7 & 30.6 & 43.5 & 0.204 \\
\ding{51} & \ding{51} & \ding{55} & 16.1 & 38.7 & 50.0 & 0.270 \\
\ding{51} & \ding{55} & \ding{51} & 19.4 & 37.1 & 51.6 & 0.298 \\
\ding{51} & \ding{51} & \ding{51} & \textbf{22.6} & \textbf{41.9} & \textbf{58.1} & \textbf{0.315} \\
\Xhline{0.8pt}
\end{tabular}}
\end{center}
\end{table}

\begin{figure}[b]
  \centering
  \centerline{\includegraphics[width=1.0\linewidth]{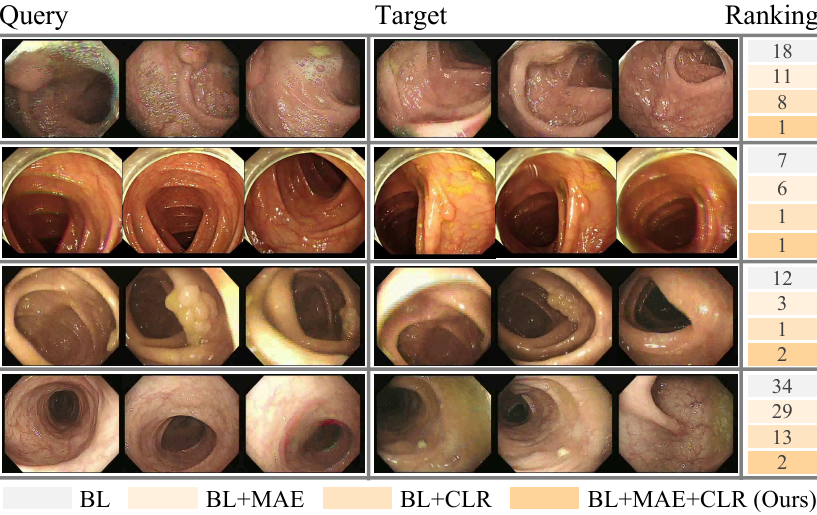}}
  \caption{Examples of ranking results between different ablation schemes of \textit{Colo-SCRL}. Our method achieves more competitive performance compared with other settings overall.}
  \label{fig4}
\end{figure}

\subsection{Ablation Study}
To verify the effectiveness of our key modules, we conduct ablation experiments in this subsection. Table \ref{tab:ablation} lists the detailed performances. Note that the baseline model only utilizes ViT pre-trained on the UCF-101 dataset as feature extractor. The best results in each column are indicated in bold. The check mark (\ding{51}) means that this module is added to the model and the cross mark (\ding{55}) means the opposite.

\noindent \textbf{Effectiveness of Video Masked Autoencoder.}
To verify the effectiveness of VideoMAE-based pre-training (MAE) to the different schemes, we conducted two groups of ablation experiments, Group I (BL with / without MAE, row 1 VS row 2 in Table \ref{tab:ablation}) and Group II (BL + CRL with / without MAE, row 3 VS row 4 in Table \ref{tab:ablation}). According to the Table \ref{tab:ablation}, using VideoMAE-based pre-training on Colo-Pair, R@1, R@5, R@10 and mAP respectively improved by \textbf{+6.4\%}, \textbf{+8.1\%}, \textbf{+6.5\%} and \textbf{+0.066} in Group I, and \textbf{+3.2\%}, \textbf{+4.8\%}, \textbf{+6.5\%}, \textbf{+0.017} in Group II, demonstrating successful general knowledge refinement during the pre-training phase.

\noindent \textbf{Effectiveness of Video Contrastive Representation Learning.} To verify the superiority of video contrastive representation learning (CRL), we conducted two groups of ablation experiments, Group III (BL with / without CRL, row 1 VS row 3 in Table \ref{tab:ablation}) and Group IV (BL + MAE with / without CRL, row 2 VS row 4 in Table \ref{tab:ablation}). As shown in Table \ref{tab:ablation}, R@1, R@5, R@10 and mAP respectively improved by \textbf{+9.7\%}, \textbf{+6.5\%}, \textbf{+8.1\%} and \textbf{+0.094} in Group III, and \textbf{+6.5\%}, \textbf{+3.2\%}, \textbf{+8.1\%}, \textbf{+0.045} in Group IV, indicating discriminative feature is captured with contrastive representation learning.

\noindent \textbf{Visualization and Qualitative Results.}
Fig. \ref{fig4} shows the qualitative results of different ablation schemes. To this end, both the quantitative and qualitative results show steady improvement and demonstrate the effectiveness of our method. Fig. \ref{fig5} also visualizes some examples of reconstruction results of a Colo-Pair video using VideoMAE.

\subsection{Discussion} We take a big step forward in colonoscopic video retrieval and obtain encouraging results on the Colo-Pair dataset. A crucial but currently less explored way to assist colonoscopy screening is to retrieve video clips of locations near the target polyp, so as to alert the doctor to the possible presence of previously found polyps in the nearby area. It is a more challenging task since it requires relying solely on nonrigid tissue information in the vicinity of the polyp. In addition, polyps retrieved in the second colonoscopy screening may not be the same as query ones in the first colonoscopy screening, which may lead to missed diagnosis and misdiagnosis to a certain extent. Polyp matching is an effective way to solve this problem. Thus, our future work will focus on polyp-free scene localization and polyp matching for better assisting CRC screening.

\begin{figure}[!t]
  \centering
  \centerline{\includegraphics[width=1.0\linewidth]{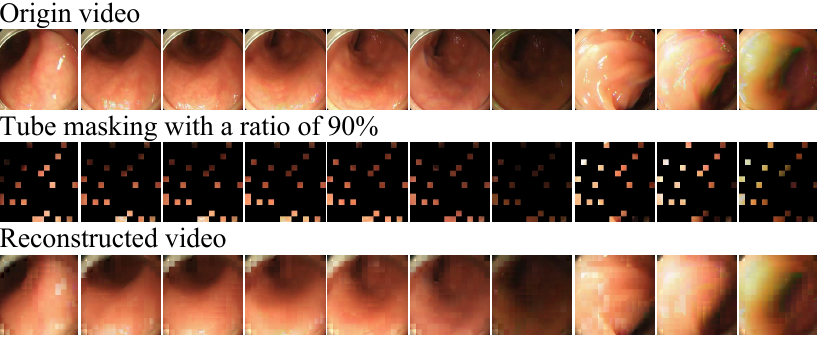}}
  \caption{VideoMAE-based reconstruction examples of Colo-Pair video.}
  \label{fig5}
\end{figure}

\section{Conclusion}
This paper investigates the problem of contrastive learning for medical video retrieval. Specifically, we manually construct a large-scale colonoscopic dataset named Colo-Pair for the first time in the medical community. On top it, a self-supervised contrastive learning representation framework is proposed for colonoscopic video retrieval. Comprehensive experiments conducted on benchmark datasets demonstrate that our method can achieve state-of-the-art retrieval performance compared with previous approaches, exhibiting strong evidence for its capability of preserving discriminative feature in representation learning. In the future, we will extend our method into other medical vision tasks, such as polyp-free scene localization and anatomical colon segment classification.

\bibliographystyle{IEEEbib}
\bibliography{icme2023}
\vspace{12pt}

\end{document}